\documentclass[9pt,journal, compsoc]{IEEEtran}

\ifCLASSOPTIONcompsoc
  \usepackage[nocompress]{cite}
\else
  \usepackage{cite}
\fi

\ifCLASSINFOpdf
   \usepackage[pdftex]{graphicx}
  \else
  \fi
\usepackage{amsmath}
\usepackage{amssymb}
\usepackage{multirow}
\usepackage{longtable}
\usepackage{eqnarray,amsmath}
\usepackage{subfig}
\usepackage{xcolor}
\usepackage{comment}
\DeclareMathOperator*{\argmin } {arg\, min}

\usepackage{algorithm2e}
\begin{document}
\title{Supervised COSMOS Autoencoder: Learning Beyond the Euclidean Loss!}

\author{Maneet Singh,~\IEEEmembership{Student Member,~IEEE,}
        Shruti Nagpal,~\IEEEmembership{Student Member,~IEEE,}
        Mayank Vatsa,~\IEEEmembership{Senior Member,~IEEE,}
        Richa Singh,~\IEEEmembership{Senior Member,~IEEE,}
        and~Afzel Noore,~\IEEEmembership{Senior Member,~IEEE}
\IEEEcompsocitemizethanks{\IEEEcompsocthanksitem M. Singh, S. Nagpal, M. Vatsa, and R. Singh are with IIIT-Delhi, India. A. Noore is with Texas A\&M University, Kingsville, USA. E-mail: \{maneets, shrutin, mayank, rsingh\}@iiitd.ac.in, afzel.noore@tamuk.edu
\protect\\ }}


\IEEEtitleabstractindextext{%

\begin{abstract}
Autoencoders are unsupervised deep learning models used for learning representations. In literature, autoencoders have shown to perform well on a variety of tasks spread across multiple domains, thereby establishing widespread applicability. Typically, an autoencoder is trained to generate a model that minimizes the reconstruction error between the input and the reconstructed output, computed in terms of the Euclidean distance. While this can be useful for applications related to unsupervised reconstruction, it may not be optimal for classification. In this paper, we propose a novel Supervised COSMOS Autoencoder which utilizes a multi-objective loss function to learn representations that simultaneously encode the (i) ``similarity" between the input and reconstructed vectors in terms of their direction, (ii) ``distribution" of pixel values of the reconstruction with respect to the input sample, while also incorporating (iii) ``discriminability" in the feature learning pipeline. The proposed autoencoder model incorporates a Cosine similarity and Mahalanobis distance based loss function, along with supervision via Mutual Information based loss. Detailed analysis of each component of the proposed model motivates its applicability for feature learning in different classification tasks. The efficacy of Supervised COSMOS autoencoder is demonstrated via extensive experimental evaluations on different image datasets. The proposed model outperforms existing algorithms on MNIST, CIFAR-10, and SVHN databases. It also yields state-of-the-art results on CelebA, LFWA, Adience, and IJB-A databases for attribute prediction and face recognition, respectively. 

\end{abstract}

}

\graphicspath{{./Images/}}

\maketitle

\IEEEdisplaynontitleabstractindextext

\IEEEpeerreviewmaketitle

\IEEEraisesectionheading{\section{Introduction}\label{sec:introduction}}

\IEEEPARstart{}{}

Traditionally, most classification tasks suffer from the inherent challenge of extracting \textit{representative} features from the given data, followed by performing effective classification. In order to learn a robust classification model, the extracted features should be invariant to modifications in the input space, capture the distinct properties of the input samples, and be representative of the data. Research in this area has been progressing over the last few decades, with developments across different spectra of hand-crafted and learning based algorithms. The past decade has specifically witnessed several advancements in this area, with large focus on deep learning techniques \cite{lecun2015deep}. Majority of the research in deep learning focuses on Convolutional Neural Network (CNN) and several advancements have been achieved with it. Additionally, other deep learning algorithms such as Autoencoder, Deep Belief Network, and Deep Boltzmann Machine have also shown promises, and we believe that increased research focus may enable future growth of these algorithms. This research builds upon this philosophy and focuses on extending the capabilities of autoencoder based representation learning.

Autoencoders are unsupervised deep learning models, utilized for learning representations of the given input data \cite{ae}.
For input data $\mathbf{X}$, the loss function of a traditional single layer autoencoder is formulated as:
\begin{equation} \label{lossAE}
\argmin_{\mathbf{W},\mathbf{W'}}  \|\mathbf{X} - \mathbf{W'}\phi(\mathbf{W X})\|^2_F  
\end{equation}
where, $\mathbf{W, W'}$ correspond to the encoding and decoding weights of the autoencoder model, respectively. $\mathbf{X}$ contains vectorized samples stacked column-wise. For example, if there are $n$ samples each with dimension $[64\times64\times3]$, $\mathbf{X}$ corresponds to a matrix of dimension $[n\times 12288]$. $\phi$ represents the activation function, which can correspond to linear (unit) or non-linear activation such as $sigmoid$, $tanh$, or $ReLU$. The model learns the representation ($\mathbf{\phi(WX)}$) such that the Euclidean distance between the reconstruction ($\mathbf{\hat{X} = W'\phi(WX)}$) and the input sample ($\mathbf{X}$) is minimized. Using the above equation, if the model learns a representation of dimension $3072$, the encoding weights ($\mathbf{W}$) have a dimension of $[12288\times3072]$, while the decoding weights ($\mathbf{W'}$) have dimension $[3072\times12288]$.  

In the literature, autoencoder and its variants have been shown to perform well on a variety of tasks such as face detection and recognition, object and speech recognition, as well as bio-medical applications \cite{majumdar2016,xu16,varAE, aeAAAI16, zhang14}. Improvements have been proposed to the autoencoder model by introducing different regularization techniques, such as $\ell_1$-norm and $\ell_{2}$-norm \cite{sparse}. These techniques are often applied on the weight matrix and result in the following loss function:
\begin{equation} \label{lossAEReg}
\argmin_{\mathbf{W},\mathbf{W'}}  \|\mathbf{X} - \mathbf{W'}\phi(\mathbf{W X})\|^2_F + \lambda R 
\end{equation}
where, $R$ corresponds to the additional regularization term, and $\lambda$ refers to the regularization parameter. One of the most popular variants of the traditional autoencoder model is the Denoising Autoencoder, which learns features that are robust to noise in the input space \cite{sdae}. Models such as the Contractive and Higher Order Contractive Autoencoder have also been proposed which learn representations robust to different variations by localizing the input space \cite{highContract, Contractive}.

\begin{table*}
\begin{center}
\caption{Brief literature review of autoencoder based formulations.}
\label{tab:auto}
\noindent\makebox[\linewidth]{
\begin{tabular}{|p{3.7cm}|p{11.5cm}|c|}
\hline
\textbf{Authors} & \textbf{Approach} & \textbf{Supervised} \\
\hline\hline
Vincent \textit{et al.} (2010) \cite{sdae} & Stacked Denoising Autoencoder (SDAE): Noise is introduced at the input layer\ & No\\
\hline
Ng  (2011) \cite{sparse} & $\ell_1$ norm is introduced in the loss function of an AE to learn sparse representations & No \\
\hline
Rifai \textit{et al.} (2011) \cite{Contractive} & Contractive AE (CAE): Added penalty term - Jacobian of input wrt hidden layer & No \\
\hline
Hinton \textit{et al.} (2011) \cite{transformingAE} & Transforming Auto-encoder: combination of proposed capsules & Yes \\
\hline
Wang \textit{et al.} (2014) \cite{generalizedAE} & Generalized AE: Incorporates the structure of the dataspace in the representation & No \\
\hline
Kingma \textit{et al.} (2014) \cite{vae} & Variational AE: Generate synthetic data by learning the training data distribution& No \\
\hline
Zhang \textit{et al.} (2015) \cite{smcae} & Stacked Multichannel AE: Learns mapping to reduce gap b/w real \& synthetic data & No \\
\hline
Gao \textit{et al.} (2015) \cite{gao2015} & Mimics SDAE - probe is noisy input, and gallery images are expected output & Yes \\
\hline
Zhuang \textit{et al.} (2015) \cite{transferAE} & Class labels are encoded in the final layer to incorporate supervision & Yes \\
\hline
Ghifary \textit{et al.} (2015) \cite{mtae} & Multi-task AE: Single representation has multiple outputs for different domains & No \\
\hline
Majumdar \textit{et al.} (2017) \cite{majumdar2016}& L-CSSE: Incorporated a group sparse regularizer to learn class-specific features & Yes \\
\hline
Meng \textit{et al.} (2017) \cite{relational} & Introduced relational term to model the relationship b/w the input data & No \\
\hline
Wang \textit{et al.} (2017) \cite{aeAAAI} & FSAE: Incorporated feature selection in AE & Yes \\
\hline
Zhang \textit{et al.} (2017) \cite{conditional} & Conditional Adversarial AE: Generate identity specific data for age variations& Yes \\
\hline
Tran \textit{et al.} (2017) \cite{cra} & Cascaded Residual AE learns difference between input data and completed data & No \\
\hline
Sethi \textit{et al.} (2018) \cite{rcodean} & R-Codean: Residual autoencoder with Cosine and Euclidean distance based loss function & No \\
\hline
Zeng \textit{et al.} (2018) \cite{coupled} & Coupled Deep AE: Learns features of LR and HR image patches, along with a mapping & Yes \\
\hline
Kodirov \textit{et al.} (2018) \cite{semantic} &Semantic AE: Additional constraint on decoder to reconstruct original visual feature  & No\\
\hline
\end{tabular}
}
\label{tab:lit}
\end{center}
\vspace{-10pt}
\end{table*}

In an attempt to encode class specific information, researchers have also proposed different autoencoder architectures by incorporating class labels during the feature learning process. Regularization techniques such as $\ell_{2,1}$-norm or group sparse regularizer introduce supervision in autoencoders \cite{anush17}. Zheng \textit{et al.} \cite{contrast} proposed the Contrastive Autoencoder, which learns representations while reducing the inter-class variations. Gao \textit{et al.} \cite{gao2015} proposed Supervised Autoencoder which utilized the class label information of the gallery and probe (same or different) to reduce the difference in representations.
Singh \textit{et al.} \cite{singh17} presented a Class Representative Autoencoder which learns features while reducing the intra-class variations and increasing the inter-class variations. To the best of our knowledge, all existing models work with a Euclidean distance based autoencoder. Only recently, Sethi \textit{et al.} \cite{rcodean} proposed a residual autoencoder which incorporates Cosine and Euclidean distance in the loss function of an autoencoder. 

A separate area of research focuses on the use of Variational Autoencoder (VAE) \cite{vae},\cite{larsen16} for the task of synthetic data generation. VAEs attempt to learn the training data distribution in order to generate synthetic data from it. While VAEs have gained significant attention over the past few years, it is important to note that it is primarily used for data generation, as opposed to learning representations for classification. Further, Hinton \textit{et al.} \cite{transformingAE} built upon the traditional autoencoder network and proposed Capsules for learning effective representations.    
Table \ref{tab:auto} summarizes some of the recently proposed autoencoders and its variants. It is important to note that while researchers have focused on learning discriminative features useful for classification using autoencoders, majority of the techniques focus on adding a penalty term along with the Euclidean distance based reconstruction error.

\subsection{Research Contributions}
In this research, keeping the goal of classification in mind, we examine the philosophy of a Euclidean distance based Autoencoder for the task of feature learning. We believe that while encoding feature vectors using the Euclidean loss generates \textit{representative} features for reconstruction, it might not result in optimal features for classification. We propose a multi-objective loss function based formulation, termed as Supervised COSine MahalanObiS (COSMOS) Autoencoder. The loss function of the proposed model aims at learning representations by encoding the (i) similarity between the input and reconstructed vectors in terms of their direction, (ii) distribution of pixel values of the reconstructed output with respect to the input sample, while incorporating (iii) discriminability in the feature learning process. This is achieved by building a model which incorporates Cosine similarity and Mahalanobis distance, along with an additional Mutual Information based penalty term for supervision. Cosine similarity is able to encode the direction variations between image vectors, while Mahalanobis distance attempts to model the pixel distributions, and Mutual Information introduces supervision. Detailed analysis of the proposed model, along with experimental results on benchmark datasets and challenging face analysis tasks further highlight the usability of the proposed Supervised COSMOS autoencoder. In the following section, each component of the proposed model is explained in detail, along with the final proposed autoencoder model. 

\begin{figure}
\begin{center}
\subfloat[(i) Euclidean Autoencoder]{\includegraphics[width= 3.1in]{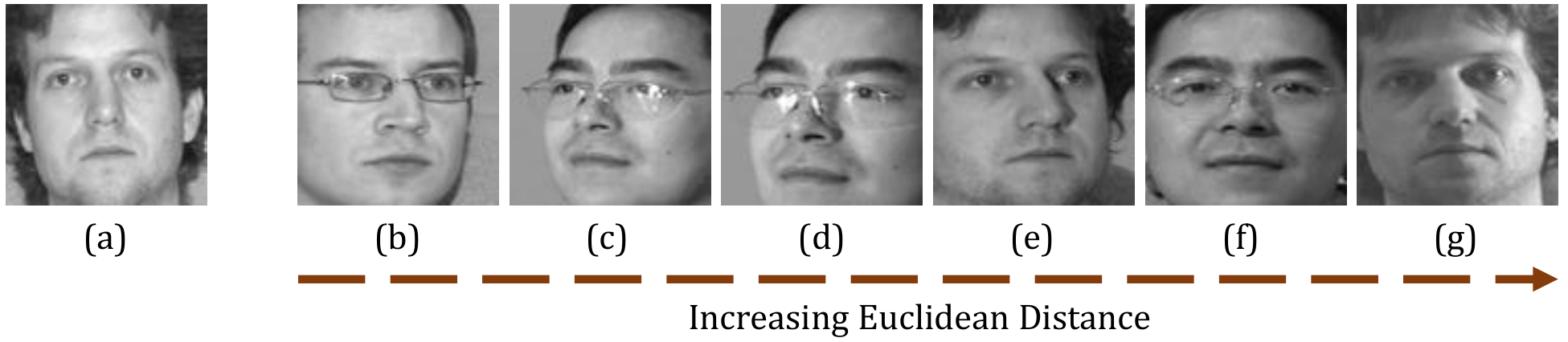}} \\
\subfloat[(ii) Proposed Supervised COSMOS Autoencoder]{\includegraphics[width= 3.1in]{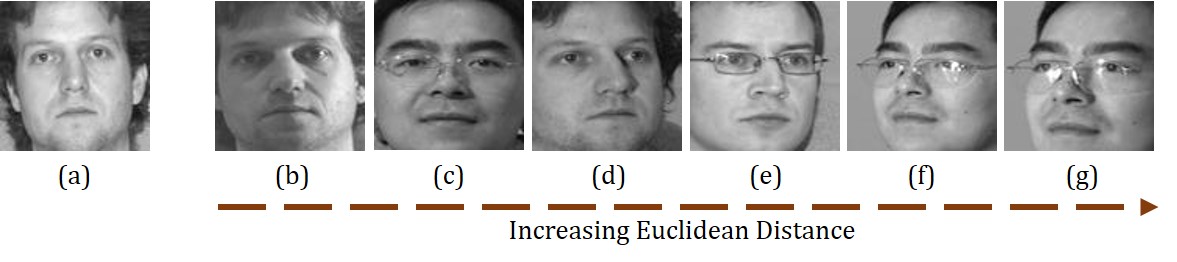}} 
\caption{Illustrating the limitations of using Euclidean distance based autoencoder, along with the advantage of the proposed Supervised COSMOS autoencoder. Images (b)-(g) are sorted by increasing distance from image (a). 
}
\label{fig:motive}
\vspace{-10pt}
\end{center}
\end{figure}

\section{Proposed Supervised COSMOS Autoencoder}

The loss function of a traditional autoencoder minimizes the mean squared error (Euclidean distance) between pixel values of the input and the reconstructed image. As shown in Fig. \ref{fig:motive} (i), this may not necessarily yield the best weight vectors to classify an image with large variations compared to the training data. 
A pre-trained autoencoder (trained on faces) is used for extracting features from these images. The Euclidean distance between the representations of each image (b)-(g) are calculated with the first image (a). The images are then sorted by increasing distance. Based on the distances calculated, it is observed that the distance between representations (of same individual) with large illumination variation is higher, as compared to features of different individuals under similar pose or illumination settings. A similar trend is observed in the image space as well. This implies that for an input image (a), a Euclidean distance based autoencoder (that is used for classification) may prefer having (b) or (c) at the reconstruction layer, i.e. images of different subjects with similar illumination, as opposed to (g) which is the same subject's image with minor illumination variation. This suggests that while Euclidean distance works well with images of similar distribution, different covariates of face recognition may affect the classification performance.

Inspired by these observations, in this research, we propose a multi-objective loss function for an autoencoder, which is able to learn representations while encoding the (i) ``direction'' variations between image vectors, (ii) ``distribution'' of pixel values, while incorporating (iii) ``supervision''. The formulation of a traditional autoencoder is modified to incorporate two different distance metrics, \textit{Cosine} and \textit{Mahalanobis}. Both these metrics are more resilient to non-identically and non-independently distributed feature vectors. This enables the feature learning model to incorporate the direction, and magnitude of the loss between the input and its reconstruction. 
Since the aim of a classification pipeline is to obtain improved classification performance, we also incorporate \textit{supervision} in the formulation of the proposed autoencoder. This is accomplished by using Mutual Information (MI) between the original class labels and predicted labels as a penalty term in the loss function. If the mutual information is high, the dependence between the two vectors is high, thus resulting in good classification accuracy. This introduces discriminability during the feature learning process. As a toy example, Fig. \ref{fig:motive} (ii) also presents the rank-list obtained by using a trained Supervised COSMOS autoencoder for feature extraction. It is motivating to observe that the proposed model encodes features invariant to changes in the input space. In contrast to the Euclidean distance based autoencoder (Fig. \ref{fig:motive} (i)), features extracted by the proposed model are able to correctly match the probe against the given gallery set. We next describe the formulation of the proposed \textit{Supervised COSMOS Autoencoder}, along with the optimization.   

\subsection{COSMOS: Autoencoder with Cosine Similarity and Mahalanobis Distance}
Cosine similarity models the similarity between two vectors in terms of the direction variations. It calculates the similarity based on the relationship of the vector values in contrast to the absolute magnitude difference between the two. Therefore, it has extensively been used in subspace learning algorithms that attempt to find vectors that best represent the given set of classes. In order to incorporate the first objective of encoding ``direction information'' we propose to utilize Cosine similarity between the input and the output in an autoencoder, i.e.:

\begin{figure}[tbp]
\begin{center}
\includegraphics[width=3in]{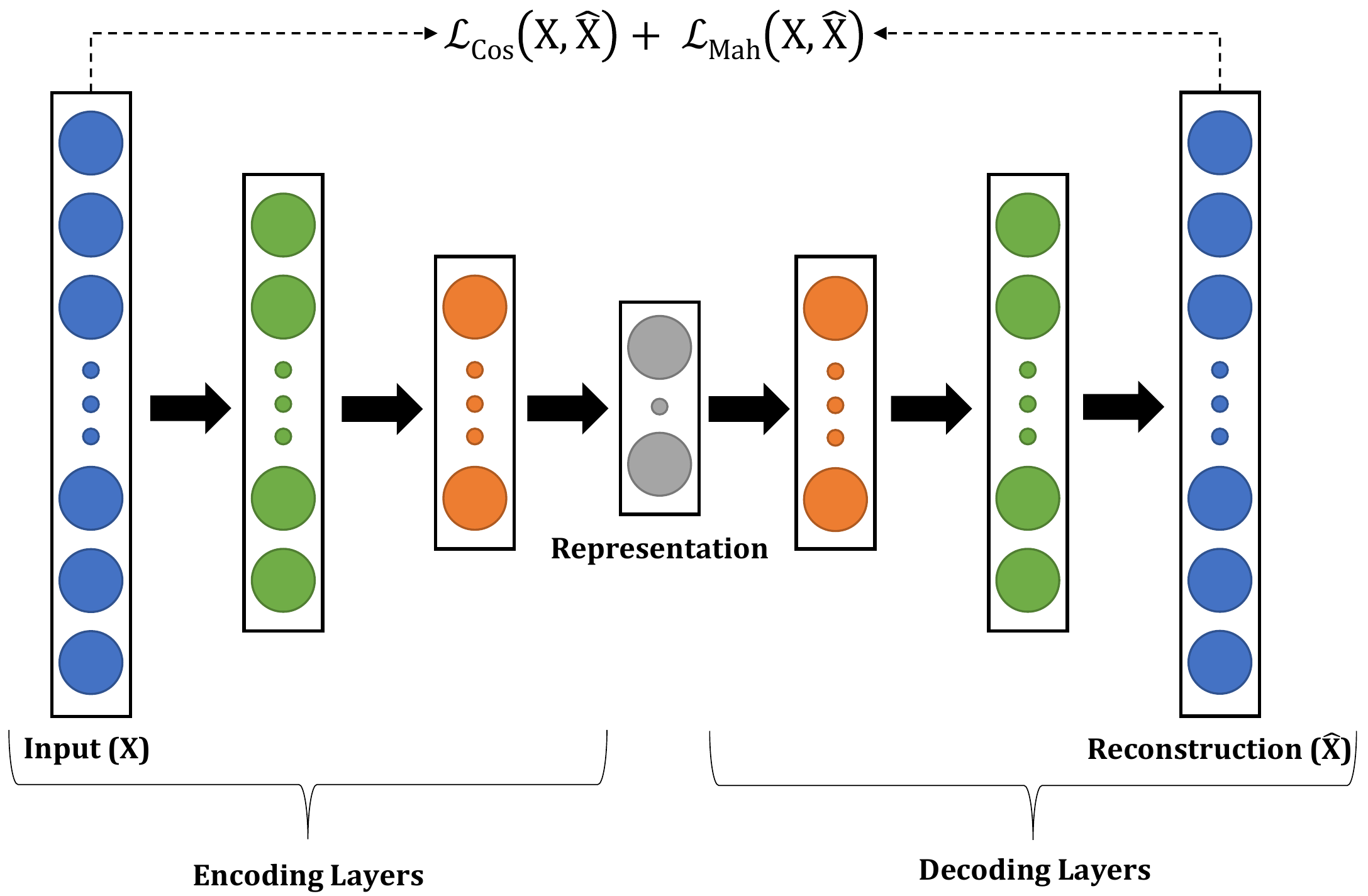}
\caption{Proposed COSMOS autoencoder with 3 hidden layers.}
\vspace{-15pt}
\label{fig:block}
\end{center}
\end{figure}

\begin{equation}
\label{eq:eq3}
\begin{split}
\mathcal{L}_{Cos}(\mathbf{X}, \mathbf{\hat{X}}) = & \|\mathbf{X} \odot (\mathbf{W'}\phi(\mathbf{W X}))\|^2 
\\ = & \frac{\mathbf{X} \cdot (\mathbf{W'}\phi(\mathbf{W X}))}{ \|\mathbf{X}\|_F^2 \times \|\mathbf{W'}\phi(\mathbf{W X})\|_F^2} 
\end{split}
\end{equation}
\noindent $\mathcal{L}_{Cos}(\mathbf{X}, \mathbf{\hat{X}})$ represents the Cosine similarity between input $\mathbf{X}$ and reconstruction $\mathbf{\hat{X}}$. $\odot$ represents the Cosine similarity operator. An autoencoder model with Cosine similarity and regularizer $R$ can be represented as:
\begin{equation} \label{eq4}
\argmin_{\mathbf{W}, \mathbf{W}'} \left( - ||\mathbf{X} \odot \mathbf{W}'\phi(\mathbf{WX})||^2 + \lambda R \right)
\end{equation}

\noindent where, $R$ is the regularizer and $\lambda$ is the regularization constant. As opposed to Euclidean distance based autoencoder, the above model does not attempt to replicate the pixel values of the input data at the reconstruction layer. On the other hand, it learns representations such that the relationship between the pixels at the reconstruction layer is similar to that of the input layer. For instance, the Cosine autoencoder would be invariant to illumination variations between the input and reconstruction. 

The second objective of modeling the distribution of pixel values of the reconstruction with respect to the input sample is achieved by utilizing the Mahalanobis distance. Mahalanobis distance accounts for the variability in the data distribution and is a unit-less scale-invariant distance metric which is used to measure the distance between two given points. For input $\mathbf{X}$ and its reconstruction $\mathbf{\hat{X}}$, it is mathematically expressed as: 
\begin{equation}
\begin{split}
\label{eq:eq41}
\mathcal{L}_{Mah}(\mathbf{X}, \mathbf{\hat{X}}) = & \|\mathbf{X} \oplus (\mathbf{W'}\phi(\mathbf{W X}))\|^2 \\
= &
 (\mathbf{X}-(\mathbf{W'}\phi(\mathbf{WX})))^T \mathbf{\mathbf{M}}(\mathbf{X}-(\mathbf{W'}\phi(\mathbf{W X})))
\end{split}
\end{equation}
\noindent where, $\mathcal{L}_{Mah}$ represents the squared Mahalanobis (pseudo) distance between $\mathbf{X}$ and $\mathbf{\hat{X}} = \mathbf{W'}\phi(\mathbf{W X})$. $\mathbf{M}$ represents a pseudo-distance matrix having the dimensions $[m\times m]$, where $m$ corresponds to the vectorized dimension of the input sample. Traditionally, in Mahalanobis distance calculations, $\mathbf{M}$ is a symmetric positive semi-definite matrix, however, for minimizing the reconstruction error of the autoencoder model, these constraints are relaxed. 
From Equation \ref{eq:eq41}, it can be seen that Euclidean distance is a special case of Mahalanobis distance, where M is an identity matrix. Thus, Mahalanobis distance is less constrained than Euclidean distance, encoding the distribution of data as well. The autoencoder formulation with Mahalanobis (pseudo) distance can be represented as:
\begin{equation} \label{eq5}
\argmin_{\mathbf{W}, \mathbf{W}', \mathbf{\mathbf{M}}} \left( ||\mathbf{X} \oplus (\mathbf{W}'\phi(\mathbf{WX}))||^2 + \lambda R \right)
\end{equation}
\noindent Minimizing the Mahalanobis distance ensures weight vectors are selected such that the distance between the input and its reconstruction is minimized when both are projected onto M. This implies that the learned representation encodes information invariant to minor manipulation of pixels such as rotation or illumination. We next combine the two objective functions and propose COSMOS autoencoder, i.e.:
\begin{equation} \label{cosmos}
\begin{gathered}
\argmin_{\mathbf{W}, \mathbf{W}', \mathbf{\mathbf{M}}, } ( - ||\mathbf{X} \odot \mathbf{\hat{X}}||^2 + ||\mathbf{X} \oplus \mathbf{\hat{X}}||^2  + \lambda R )
\end{gathered}
\end{equation}
Fig. \ref{fig:block} presents a pictorial representation of the COSMOS autoencoder. Cosine similarity and Mahalanobis distance based loss function facilitates learning of robust representations, however, it does not introduce discriminability with respect to the class labels. This is incorporated with the help of a Mutual Information based penalty term.

\subsection{Supervised COSMOS: Incorporating Supervision with Mutual Information}
The final objective of learning discriminative features is achieved via Mutual Information. Mutual Information has successfully been used in several image processing tasks including image registration. Recently, it has been incorporated to encode supervision in the feature extraction process \cite{mi}. This leads to learning discriminative features which enhances the classification performance. Mutual Information (MI) is defined as:
\begin{equation} \label{eq61}
\begin{gathered}
MI(\mathbf{Y}_\mathbf{P}, \mathbf{Y}_\mathbf{L}) = p(\mathbf{Y}_\mathbf{P}, \mathbf{Y}_\mathbf{L}) log\left(\frac{p(\mathbf{Y}_\mathbf{P}, \mathbf{Y}_\mathbf{L})}{p(\mathbf{Y}_\mathbf{P})p(\mathbf{Y}_\mathbf{L})}\right)
\end{gathered}
\end{equation}
\noindent where, $\mathbf{Y}_\mathbf{P}$ represents the predicted label and $\mathbf{Y}_\mathbf{L}$ is the ground truth label of the input data, and $p(x)$ is the probability of x. We propose to incorporate mutual information as a penalty term to introduce supervision in the autoencoder model. A traditional autoencoder with mutual information based loss function can be represented as: 
\begin{equation} \label{eq6}
\begin{gathered}
\argmin_{\mathbf{W}, \mathbf{\mathbf{W'}}, \mathbf{ \omega} } \left( ||\mathbf{X} - \mathbf{W}'\phi(\mathbf{WX})||^2_F -  \lambda_1 MI(\mathbf{Y}_\mathbf{P}, \mathbf{Y}_\mathbf{L}; \omega) +  \lambda_2 R \right)
\end{gathered}
\end{equation}
\noindent where, $\omega$ is the weight of the mutual information based classifier $\left( \sum \omega^T\phi(\mathbf{WX}) \right)$. Mutual information between the ground truth label and the predicted label is encoded as a supervised regularizer. Since mutual information is a similarity term, it is added in the loss function with a negative sign. $\lambda_1$ and $\lambda_2$ are the regularization constants.

\begin{figure*}
\begin{center}
\includegraphics[width=5.2in]{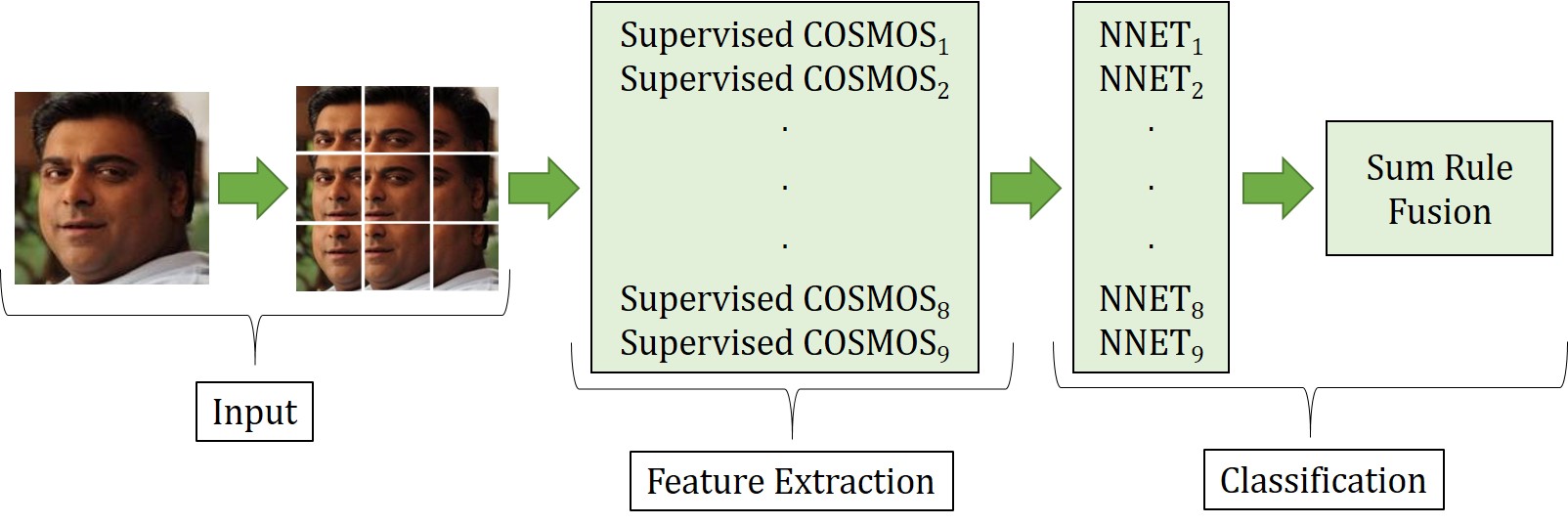}
\caption{Illustrating the steps involved for utilizing Supervised COSMOS for classification.}
\vspace{-10pt}
\label{fig:pipeline}
\end{center}
\end{figure*}

Similarly, we incorporate supervision in the proposed COSMOS autoencoder and the loss function of a single layer Supervised COSMOS Autoencoder can be written as:
\begin{equation} \label{eq6}
\begin{gathered}
\argmin_{\mathbf{W}, \mathbf{W}', \mathbf{\mathbf{M}}, \mathbf{ \omega} } \left( - ||\mathbf{X} \odot \mathbf{W}'\phi(\mathbf{WX})||^2 + ||\mathbf{X} \oplus \mathbf{W}'\phi(\mathbf{WX})||^2 \right.  \\
 \left. -  \lambda_1 MI(\mathbf{Y}_\mathbf{P}, \mathbf{Y}_\mathbf{L}; \omega) +  \lambda_2 R \right)
\end{gathered}
\end{equation}
where, $\lambda_1$ and $\lambda_2$ are the regularization constants. Thus, the proposed supervised COSMOS autoencoder builds over a traditional autoencoder by using a multi-objective loss function. It combines the Cosine similarity and Mahalanobis distance along with Mutual Information based supervision loss for learning robust features for classification.

\subsection{Optimization and Learning Parameters}
In the above mentioned formulation, the encoding and decoding weights are assumed to be tied, i.e., $\mathbf{W'} = \mathbf{W}^T$. The supervised layers of COSMOS are optimized using the alternating minimization approach \cite{am}. It is a well established approach for the minimization of a function over multiple parameters. For the $k^{th}$ iteration, the optimizations are as follows: 

\noindent \textbf{Step 1: Optimizing weight of COSMOS ($\mathbf{W}$)}:
\begin{equation} \label{eq6}
\begin{gathered}
\mathbf{W}_k \leftarrow \argmin_{\mathbf{W}} \ \mathcal{L}_{Mah}(\mathbf{X}, {\mathbf{\hat{X}}}) + \mathcal{L}_{Cos}(\mathbf{X}, {\mathbf{\hat{X}}}) \\
- \lambda_1 MI(\mathbf{Y}_\mathbf{P}, \mathbf{Y}_\mathbf{L}; \omega) + \lambda_2R
\end{gathered}
\end{equation}
\noindent \textbf{Step 2: Optimizing pseudo-covariance matrix ($\mathbf{M}$)}:
\begin{equation} \label{eq6}
\begin{gathered}
\mathbf{M}_k \leftarrow \argmin_{\mathbf{\mathbf{M}}} \ \mathcal{L}_{Mah}(\mathbf{X}, {\mathbf{\hat{X}}})
\end{gathered}
\end{equation}
\noindent \textbf{Step 3: Optimizing Mutual Information based Classifier ($\omega$):}
\begin{equation} \label{eq6}
\begin{gathered}
\mathbf{\omega_{(k)}} \leftarrow \argmin_{\omega} - \lambda_1 MI(\mathbf{Y}_\mathbf{P}, \mathbf{Y}_\mathbf{L}; \omega)
\end{gathered}
\end{equation}

The above three steps are repeated iteratively until maximum iterations are reached or model converges. $ReLU$ activation is applied on each layer and dropout is used as a regularizer. The values of regularization constants are computed experimentally by performing a grid search. In order to prevent the problem of vanishing gradients, skip connections \cite{resNet} are added in the proposed Supervised COSMOS Autoencoder. A connection is added between each alternate encoding layer which facilitates gradient flow at the time of feature learning. 

\subsection{Object/Face Classification via Supervised COSMOS}
The proposed formulation of supervised COSMOS autoencoder is applied in object and face classification applications. Fig. \ref{fig:pipeline} illustrates the pipeline adopted for the same. As shown in the image, the input image is tessellated into nine overlapping patches which are provided as input to the Supervised COSMOS autoencoder to learn discriminative features. This is done in order to encode local features. The learned features are then classified using a 2-layer Neural Network of dimension [$\frac{n}{2}$, $\frac{n}{4}$], where $n$ is the input feature size. Results from each local level are then combined using sum rule fusion. The regularization constants are updated adaptively. The proposed model is implemented in Theano using Adam optimization. The model is trained on a workstation with Intel Xeon 2.6 GHz processor with 64 GB RAM, and NVIDIA K40 GPU. 

\begin{table*}
\begin{center}
\caption{Details of datasets used in this research along with the architectural details of the proposed model.}
\begin{tabular}{|l|c|c|c|c|c|}
\hline
\textbf{Database} 	& \textbf{Classes}	& \textbf{Image Size}	& \textbf{Patch Size} & \textbf{Total Images} & \textbf{Architecture of the proposed model} \\\hline
MNIST		&10		&  $28 \times 28$  & $14 \times 14$ & 70,000	& [196; 150; 100; 100; 50]	\\
\hline
CIFAR-10		& 10		& \multirow{2}{*}{$32 \times 32 \times 3$} & \multirow{2}{*}{$16 \times 16 \times 3$} & 60,000	& [768; 600; 500; 400; 300; 200]	\\
\cline{1-2}
\cline{5-6}
SVHN		& 10		&  & & 99,289	& [768; 600;  500; 400; 300; 200; 100]	\\
\hline
CelebA		& 2		& \multirow{6}{*}{$64 \times 64 \times 3$}	& \multirow{6}{*}{$32 \times 32\times 3$}	& 200,000	& [3072; 3000; 2500; 2500; 2000; 1500; 1000; 500]	 \\ 
\cline{1-2}
\cline{5-6}
LFWA		&2		&  & & 13,233	& [3072; 3000; 2500; 2500; 2000; 1500; 1000; 500]	\\
\cline{1-2}
\cline{5-6}

Adience - Age	& 8		&  & &17,643	& [3072; 3000; 2500; 2000; 1000]	\\
\cline{1-2}
\cline{5-6}

Adience - Gender &2		&  & &19,487 	&  [3072; 3000; 2500; 2000; 1000]	\\
\cline{1-2}
\cline{5-6}

IJB-A - Identification	& 500		&  &&  5,712 & [3072; 3000; 2500; 2500; 2000; 1500; 1000; 500]	\\
\hline

\end{tabular}
\label{tab:stats}
\end{center}
\end{table*}

\section{Datasets and Experimental Protocols}
Performance of the proposed supervised COSMOS autoencoder based framework is demonstrated on seven benchmark datasets. Details regarding each are provided below:

\noindent \textbf{MNIST Dataset} \cite{mnist} has images of handwritten digits - 0 to 9, with dimensions $28\times28$. The training data contains 60,000 images pertaining to all 10 classes, whereas the test set comprises of 10,000 images. Both the training and testing sets contain equal samples from all classes. 

\noindent \textbf{CIFAR-10 Dataset} \cite{cifar} is a large image dataset of different object categories having dimensions $32\times32\times3$. It consists of 60,000 RGB images corresponding to 10 different classes. 
The dataset is divided into training and testing partitions having 50,000 and 10,000 images, respectively. Equal number of samples from each class are ensured in the training and testing sets. 

\noindent \textbf{The Street View House Numbers (SVHN) Dataset} \cite{svhn} is a real world dataset of RGB images of dimension $32\times32\times3$. It contains over 600,000 images of house numbers obtained from Google Street View images. It contains images of 10 classes - 0 to 9, which are centered around a single character. The database contains 73,257 digits for training, 26,032 digits for testing, and an additional 531,131 digits as unsupervised training set.

\noindent \textbf{CelebA Dataset} \cite{liu2015deep} is a large scale face attribute dataset containing 20 images per subject for 10,000 subjects. Each image is annotated with 40 attributes and five landmark points. The images have large pose variation and background clutter making the data challenging. The results are reported on the pre-defined protocol for attribute prediction. 

\noindent \textbf{Labeled Faces in the Wild Attributes (LFWA) Dataset} \cite{liu2015deep} consists of 13,233 images of 5,749 subjects. The dataset is created by labeling attributes in images of LFW dataset. Similar to CelebA dataset, this dataset is also used for the task of attribute prediction for the 40 attributes annotated in each image. 

\noindent \textbf{Adience Dataset} \cite{adience} contains 26,580 face images pertaining to 2,284 individuals. The images contain several variations across appearance, noise, pose, lighting, and capture devices. This dataset has primarily been used for predicting age and gender from face images. It contains labels pertaining to male and female, and eight different age groups. Pre-defined protocol for five fold cross-validation specifying the training and testing partitions has also been provided. 

\noindent \textbf{IJB-A Dataset} \cite{ijba} contains 5,712 face images and 2,085 videos of 500 individuals. The images are captured with different devices in varied environment and pose variations. The pre-defined face identification protocol is used in the experiments. 

Table \ref{tab:stats} summarizes the dataset details as well as the architecture of the proposed model. Experimental evaluation is performed using the pre-defined protocols pertaining to each dataset. All protocols ensure disjoint training and testing splits.

\begin{table}[t]
\centering
\caption{Comparison with state-of-the-art results on benchmark MNIST, CIFAR-10, and SVHN datasets.}
\label{tab:imageres}
\begin{tabular}{|l|l|c|c|c|}
\hline
& \multirow{2}{*}{\textbf{Algorithm}} & \multicolumn{3}{c|}{\textbf{Classification Error (\%)}} \\
\cline{3-5}
& & \textbf{MNIST} & \textbf{CIFAR-10}  & \textbf{SVHN} \\ 
\hline 
\hline
\multirow{4}{*}{\rotatebox[origin=c]{90}{AE}} & WTA AE \cite{wtaAE} & 0.48 & 19.90 & 6.90 \\
\cline{2-5}
& Adversarial AE \cite{aae} & 0.85 & - & - \\
\cline{2-5}
& Self-Paced AE \cite{spae} & 3.32 & - & - \\
\cline{2-5}
& GSAE \cite{anush17} & 1.10 & 22.6 & 7.6 \\
\hline
\multirow{10}{*}{\rotatebox[origin=c]{90}{CNN}} & DropConnect \cite{dropconnect} & \textbf{0.21}  & 9.32  & 1.94 \\
\cline{2-5}
& MCDNN \cite{mcdnn} & \textbf{0.23} & 11.21 & - \\
\cline{2-5}
& Gen. Pooling CNN \cite{genPoolCnn} & 0.31 & 7.62 & \textbf{1.69}\\
\cline{2-5}
& RCNN \cite{recCnn} & 0.31 & 8.69 & \textbf{1.77}\\
\cline{2-5}
& MIM \cite{pieceWise} & 0.31 & 8.52 & 1.97\\
\cline{2-5}
& FitNet \cite{goodInit} & 0.38  & 6.06 & - \\
\cline{2-5}
& Tuned CNN \cite{boDnn} & -  & 6.37 & - \\
\cline{2-5}
& ResNet \cite{resNet} & -  & 6.43 & - \\
\cline{2-5}
& Wide-Resnet \cite{wide} & -  & \textbf{4.17} & 1.64 \\
\cline{2-5}
& DensetNet \cite{denseNet} & - & \textbf{5.19} & 1.59 \\
\hline
& CapsNet \cite{capsule} & 0.25 & 10.60 & - \\
\hline
& \textbf{Proposed Framework} & \textbf{0.21} & \textbf{5.35} & \textbf{1.08} \\
\hline
\end{tabular}
\vspace{-10pt}
\end{table}

\begin{table}
\centering
\caption{Comparison with existing algorithms on CelebA and LFWA datasets. The reported accuracy is the mean classification accuracy obtained over all the attributes.}
\label{tab:lfw}
\begin{tabular}{|l|l|c|c|}
\hline
& \textbf{Architecture} & \textbf{CelebA} & \textbf{LFWA}         \\     
\hline
\hline
\multirow{2}{*}{\rotatebox[origin=c]{90}{AE}} & Sethi \textit{et al.} \cite{rcodean} &90.14 & 84.80 \\ 
\cline{2-4}
& Hou \textit{et al.} \cite{houVAE} & 88.73 & - \\
\hline
\multirow{9}{*}{\rotatebox[origin=c]{90}{CNN}} & Wang \textit{et al.} \cite{wang2016walk}& 88.00 & 87.00  \\     
\cline{2-4}
& Zhong \textit{et al.} \cite{zhong2016leveraging} & 89.80 & 85.90 \\     
\cline{2-4}
& Rozsa \textit{et al.} \cite{rozsa2016facial} & 90.80 & - \\     
\cline{2-4}
& Rudd \textit{et al.} \cite{rudd2016moon} & 90.94 & - \\ 
\cline{2-4}
& Hand and Chellappa \textit{et al.} \cite{hand17aaai} & 91.26 & 86.30 \\ 
\cline{2-4}
& Kalayeh \textit{et al.} \cite{kalayeh_17_cvpr} & 91.80 & 87.13 \\ 
\cline{2-4}
& He \textit{et al.} \cite{ijcai18He} & 91.81 & 85.28 \\
\cline{2-4}
& Wang \textit{et al.} \cite{wang17} & 92.00 & - \\ 
\cline{2-4}
& Han \textit{et al.} \cite{han17} & 93.00 & 86.00 \\ 
\hline
& \textbf{Proposed Framework} & \textbf{94.14}    & \textbf{88.26} \\ 
\hline
\end{tabular}
\end{table}

\section{Results and Analysis}
The proposed Supervised COSMOS Autoencoder framework has been evaluated on three tasks: image classification, attribute prediction, and face recognition. Experiments have been performed on the benchmark MNIST, CIFAR-10, SVHN, CelebA, LFWA, Adience, and IJB-A datasets. Comparison has been performed with state-of-the-art algorithms, and other existing deep learning models. This is followed by an ablation study on the proposed framework, in order to understand the effect of each component. The following subsections discuss the results and observations across the experiments.


\subsection{Classification Performance}
Tables \ref{tab:imageres} - \ref{tab:iden} present the classification performance of the proposed supervised COSMOS framework for the three tasks of image classification, attribute prediction, and face recognition. Comparison has also been performed with the current state-of-the-art techniques and other deep learning algorithms. 

\textbf{Image Classification:} Table \ref{tab:imageres} presents comparison of the proposed Supervised COSMOS model with state-of-the-art (peer-reviewed) results reported on MNIST, CIFAR-10, and SVHN datasets. In case an algorithm does not show results on a particular dataset, it is represented as a `-'. It can be observed that the proposed model achieves improved or comparable performance on all three datasets as compared to state-of-the-art and other existing CNN based algorithms. For MNIST, Supervised COSMOS achieves an error of 0.21\%, which is equivalent to the best reported result \cite{dropconnect}. On the SVHN dataset, the proposed model achieves a classification error of 1.08\%, thus reporting an improvement over the current state-of-the-art result. On the other hand, it achieves an error of 5.35\% on the CIFAR-10 dataset, and is among the top-3 performing models on this dataset.

\begin{figure*}[t]
\begin{center}
\subfloat[(a) Eye Glasses]{\includegraphics[width = 6in]{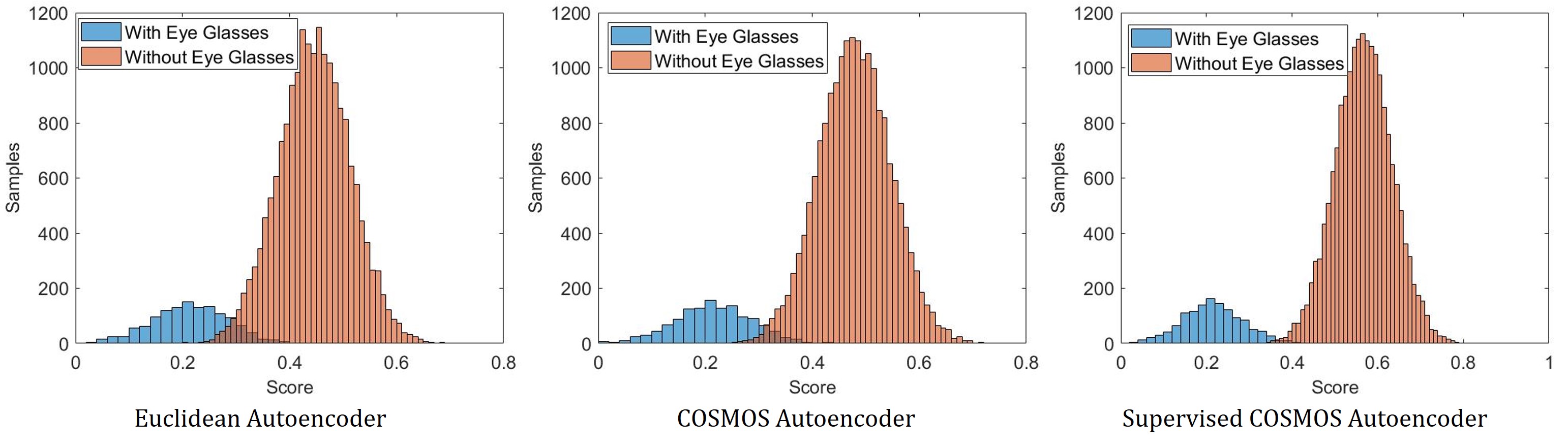}} 
\vspace{-5pt}
\\
\subfloat[(b) Oval Face]{\includegraphics[width=6in]{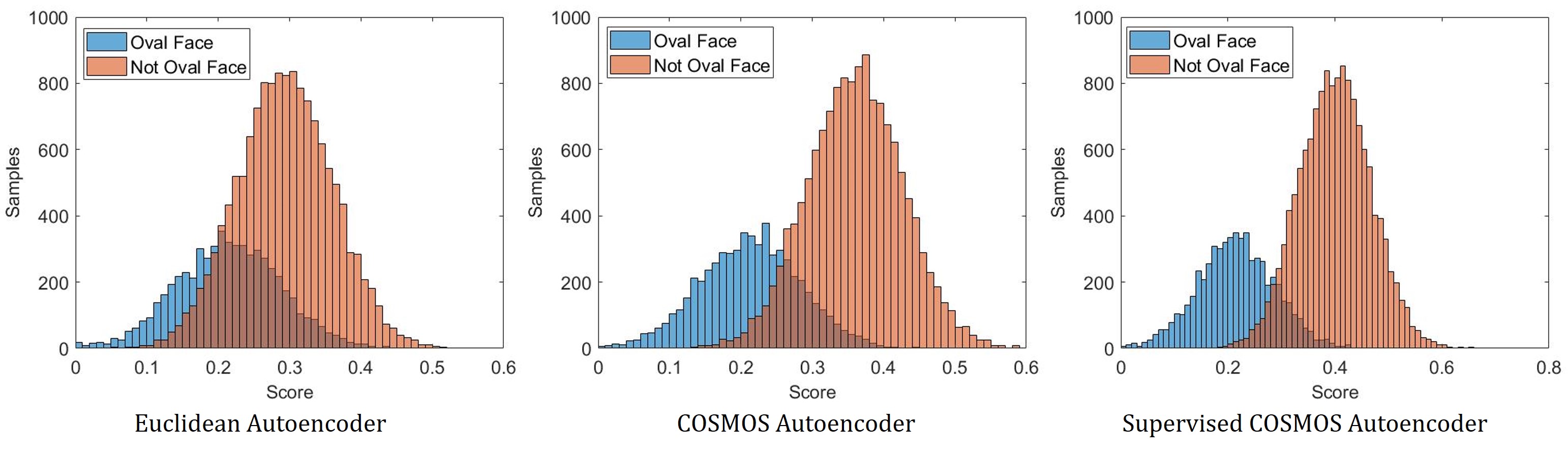}} 
\vspace{-5pt}
\caption{Score distribution of CelebA test samples for the best and worst performing attribute, \textit{Eye Glasses} and \textit{Oval Face}. Comparison can be performed across the traditional Euclidean distance based autoencoder, COSMOS, and Supervised COSMOS autoencoder.}
\vspace{-10pt}
\label{fig:scores}
\end{center}
\end{figure*}

\begin{table}[]
\centering
\caption{Classification accuracies (\%) of existing algorithms and the proposed model on Adience Dataset.}
\label{tab:adience}
\begin{tabular}{|l|l|c|c|}
\hline
& \textbf{Algorithm} & \textbf{Gender} & \textbf{Age} \\
\hline
\hline
\multirow{6}{*}{\rotatebox[origin=c]{90}{CNN}} & Levi and Hassner \cite{leviCvpr} & 86.8 $\pm$ 1.4 & 50.7 $\pm$ 5.1 \\
\cline{2-4}
& DEX \cite{dex} & - & 64.0 $\pm$ 4.2 \\
\cline{2-4}
& CNN + SVM \cite{cnnSvm} & 87.2 $\pm$ 0.7 & - \\
\cline{2-4}
& Ft-VGG-Face + SVM \cite{transfer} & 92.0 & 57.9 \\
\cline{2-4}
& FDAR-NET \cite{fdar} & 92.5 & \textbf{80.5} \\
\cline{2-4}
& VGG-Face + Attention \cite{pr17} & 93.0 $\pm$ 1.8 & 61.8 $\pm$ 2.1 \\ 
\hline 
\hline
& \textbf{Proposed Framework} & \textbf{95.07 $\pm$ 0.2} & \textbf{77.98 $\pm$ 0.6} \\
\hline
\end{tabular}
\end{table}

\begin{table}
\begin{center}
\caption{Confusion matrix of Supervised COSMOS Autoencoder on the Adience dataset for gender classification.}
\label{tab:confGender}
\begin{tabular}{ | p{0.3cm} || c | c|c|  } 
\hline
& \multicolumn{3}{c|}{\textbf{Predicted}} \\ 
\hline
\hline
\multirow{3}{*}{\rotatebox[origin=c]{90}{\textbf{Actual}}}&  & Male & Female \\
\cline{2-4}
& Male & 94.68\% & 5.32\% \\
\cline{2-4}
& Female & 4.56\% & 95.44\%  \\
\hline
\end{tabular}
\end{center}
\vspace{-15pt}
\end{table}

\begin{table}[t]
\centering
\caption{Identification results (\%) on the IJB-A face dataset.}
\label{tab:iden}
\begin{tabular}{|l|l|c|c|}
\hline
& \textbf{Algorithm} & \textbf{Rank-1} & \textbf{Rank-10}\\
\hline
\hline
\multirow{5}{*}{\rotatebox[origin=c]{90}{CNN}} & DCNN$_{Manual}$+Metric \cite{dcnn} &  0.852 $\pm$ 0.018 & 0.954 $\pm$ 0.007   \\
\cline{2-4}
& NAN \cite{hua-2016}				& 0.958 $\pm$ 0.005   	& 0.986 $\pm$ 0.003 \\
\cline{2-4}
& TDFF [36]+TPE \cite{tdff-tpe} 				&  0.964 $\pm$ 0.007 	& 0.992 $\pm$ 0.003 \\
\cline{2-4}
& L2-S (RX101)  \cite{chellappa-l2} 	& \textbf{0.977 $\pm$ 0.005}	& \textbf{0.993 $\pm$ 0.002} \\
\cline{2-4}
& VGGFace2 \cite{vggface2} & \textbf{0.982$\pm$0.004}& \textbf{0.994$\pm$0.001}\\
\hline
\hline
& \textbf{Proposed Framework} 				& \textbf{0.976 $\pm$ 0.003} & \textbf{0.992 $\pm$ 0.002} \\\hline
\end{tabular}
\vspace{-10pt}
\end{table}

\textbf{Attribute Prediction:} Classification accuracies of the proposed model for CelebA and LFWA datasets are reported in Table \ref{tab:lfw}. Pre-defined protocols are used to perform attribute prediction for the 40 annotated attributes in CelebA and LFWA datasets, and gender and age classification on the Adience dataset. In literature, Han \textit{et al.} \cite{han17} obtain the best performance of 93.00\% on the CelebA dataset. The proposed approach yields an improvement of around 1.1\% on it, resulting in 94.14\%. Similarly on the LFWA dataset, the proposed approach obtains a mean accuracy of 88.26\%, demonstrating improvement over the existing state-of-the-art results presented by Kalayeh \textit{et al.} \cite{kalayeh_17_cvpr}. These results illustrate the efficacy of the proposed model on large datasets, thereby encouraging the use of proposed Supervised COSMOS framework. 
 
Table \ref{tab:adience} presents the gender and age classification accuracies on the Adience dataset. FDAR-NET \cite{fdar} yields the best accuracy of 92.5\% and 80.5\% for gender and age classification, respectively. It can be observed that the proposed Supervised COSMOS autoencoder achieves a gender classification accuracy of 95.07\%, and an age classification accuracy of 77.98\%. The model improves the current state-of-the-art by 2.57\% for gender classification, while achieving second best results for age classification. The proposed Supervised COSMOS autoencoder also presents a reduced standard deviation across the five folds for both tasks. This shows that the model learns robust features for classification. Table \ref{tab:confGender} presents the confusion matrix obtained for gender classification. It can be observed that the proposed model performs well on both the classes, without being biased towards any one class. 

\textbf{Face Recognition: }The proposed supervised COSMOS framework is also evaluated on the IJB-A dataset for the task of face recognition. IJB-A dataset is a part of the IARPA's JANUS project, and is one of the most challenging face databases. As per the standard practice, Table \ref{tab:iden} presents the rank-1 and rank-10 identification accuracies on the IJB-A dataset. It can be observed that the proposed Supervised COSMOS autoencoder achieves among the best performing results on IJB-A dataset. As compared to existing architectures on both the ranks, the superior performance along with the low values of standard deviation obtained across different folds further promote the usage of the proposed model for face identification tasks.

\begin{table*}[]
\begin{center}
\caption{Ablation study on the proposed Supervised COSMOS autoencoder for SVHN and CelebA datasets.} 
\begin{tabular}{|c|c|c|c||c|c|c|}
\hline
\textbf{Effect of} & \multicolumn{3}{c||}{\textbf{CelebA}} & \multicolumn{3}{c|}{\textbf{SVHN}} \\
\hline
\hline
\textbf{Distance} & Euclidean & Cosine & Mahalanobis & Euclidean & Cosine & Mahalanobis\\
\cline{2-7}
\textbf{Metric} & 85.49 & 85.58 & \textbf{85.91} & 89.68 & 91.12 & \textbf{91.39} \\
\hline
\textbf{Supervision} & Euclidean+MI & Cosine+MI & Mahalanobis+MI & Euclidean+MI & Cosine+MI & Mahalanobis+MI\\
\cline{2-7}
\textbf{via MI} & 86.17	& 86.51 & \textbf{88.06} & 91.04 & 90.57 & \textbf{92.78} \\
\hline
\textbf{Distance Metric} & {Euc. + Cos} & {Euc. + Maha.} &
{Cos + Maha.} &
{Euc. + Cos} & {Euc. + Maha.} &
{Cos + Maha.} \\
\cline{2-7}
\textbf{Combination} & {85.74} & {86.88} &
\textbf{88.92} &
{91.44} & {90.92} &
\textbf{93.23} \\
\hline
\end{tabular}
\label{tab:compon}
\end{center}
\vspace{-15pt}
\end{table*}

\textbf{Comparison with Other Deep Learning Algorithms:} Tables \ref{tab:imageres} - \ref{tab:iden} can be analyzed to compare the performance of the proposed Supervised COSMOS Autoencoder framework with other existing deep learning techniques, specifically, autoencoder and convolutional neural network (CNN) based models. It is interesting to note that most of the top performing algorithms incorporate CNNs in their classification pipeline. The proposed technique is among the few autoencoder based frameworks which achieves improved or comparable performance to existing CNN models. It is our belief that the incorporation of supervision during the training of the proposed supervised COSMOS facilitates learning of \textit{discriminative} yet \textit{representative} features. The class specific characteristics encoded at the feature level are further accentuated while learning a classifier, thereby resulting in improved performance.

\subsection{Ablation Study on Supervised COSMOS Framework}
The proposed Supervised COSMOS Autoencoder is formulated using a multi-objective loss function combining Cosine similarity, Mahalanobis distance, and Mutual Information based supervision. In order to understand the effect of each objective function and their various combinations, ablation study has been performed with the CelebA and SVHN datasets. Table \ref{tab:compon} presents the performance of different components of the proposed framework. 

\textbf{Effect of Distance Metric:} The first set of experiments is performed to evaluate autoencoder models built using different distance metrics i.e. Euclidean distance, Cosine similarity, and Mahalanobis distance (Equations \ref{lossAE}, \ref{eq4}, \ref{eq5}). 
It is observed that Cosine similarity and Mahalanobis distance based autoencoders yield improved classification performance as compared to the traditional Euclidean distance based autoencoder. This can be attributed to the fact that while autoencoders with Euclidean distance loss function attempt to replicate the input at the reconstruction layer, Cosine similarity and Mahalanobis distance based loss functions focus on features that are invariant to minute rotation and illumination variations. This affirms our hypothesis that Euclidean distance based autoencoders might not be best suited for classification tasks. 

\textbf{Effect of Supervision via MI:} The next set of experiments analyze the effect of Mutual Information (MI) based supervision (Eq. \ref{eq61}) in the autoencoder model. MI based penalty term is added in the autoencoder formulation built using Euclidean distance, Cosine similarity, and Mahalanobis distance, independently (Table \ref{tab:compon}). The addition of MI based supervision leads to an improvement of $0.6 - 2.1\%$, except with Cosine similarity, where the accuracy on SVHN reduces slightly. This strengthens our claim that incorporating MI based supervision during feature learning helps improve the classification performance, by facilitating learning of discriminative representations.

\begin{table}[]
\centering
\caption{Accuracies (\%) obtained with varying number of layers of COSMOS model.}
\label{tab:depth}
\begin{tabular}{|c|c|c|c|c|c|c|}
\hline
\textbf{No of Layers} &\textbf{3}&\textbf{4}&\textbf{5}&\textbf{6}&\textbf{7}&\textbf{8} \\
\hline\hline
CelebA	&  85.71 &  89.33 &  92.46 &  93.98 &  \textbf{94.14} &  93.76    \\\hline
SVHN 	& 91.05		& 94.28 	&  96.31 & \textbf{98.92} & 98.51 & 98.03 \\
\hline
\end{tabular}
\vspace{-10pt}
\end{table}

\begin{figure} [t]
\begin{center}
\includegraphics[width=2.8in]{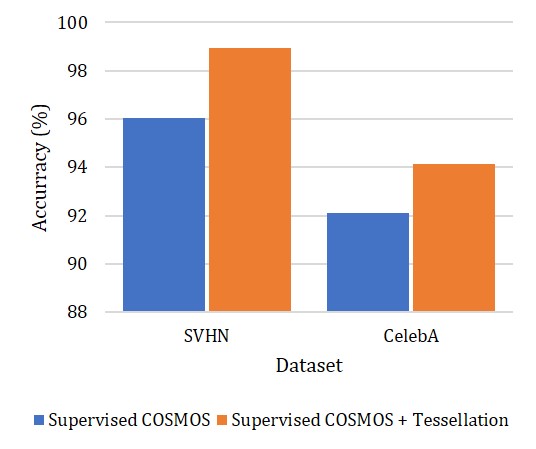}
\vspace{-10pt}
\caption{Classification performance variation before and after incorporating tessellation in the proposed framework.}
\label{fig:bar}
\end{center}
\vspace{-15pt}
\end{figure}

\textbf{Effect of Distance Metric Combination:} The third set of experiments enable us to understand the effect of combination of distance metrics for the loss function of an autoencoder. It can be observed that the autoencoder utilizing a Cosine similarity and Mahalanobis distance (COSMOS) based loss function outperforms other combinations, as well as individual loss functions for both the datasets. The COSMOS autoencoder yields $1.8-3.2\%$ higher classification accuracy than other combinations on CelebA and SVHN respectively. This also affirms our hypothesis that ``direction'' and ``distribution'' information can jointly help extract better features. Fig. \ref{fig:scores} presents the score distributions of the best and worst performing attributes, \textit{Eye Glasses} and \textit{Oval Face} of the CelebA dataset. Plots for a specific attribute can be analyzed in order to observe the progressive improvement of distributions. For the \textit{Eye Glasses} attribute, classification via the Euclidean distance based autoencoder results in a minor overlap of scores between the two classes, which is almost eliminated with the proposed supervised COSMOS autoencoder. A similar trend is observed for the \textit{Oval Face} attribute, where the traditional autoencoder suffers from a large overlap, which is significantly reduced with the proposed supervised COSMOS model.  

\begin{figure} [t]
\begin{center}
\includegraphics[width=2.8in]{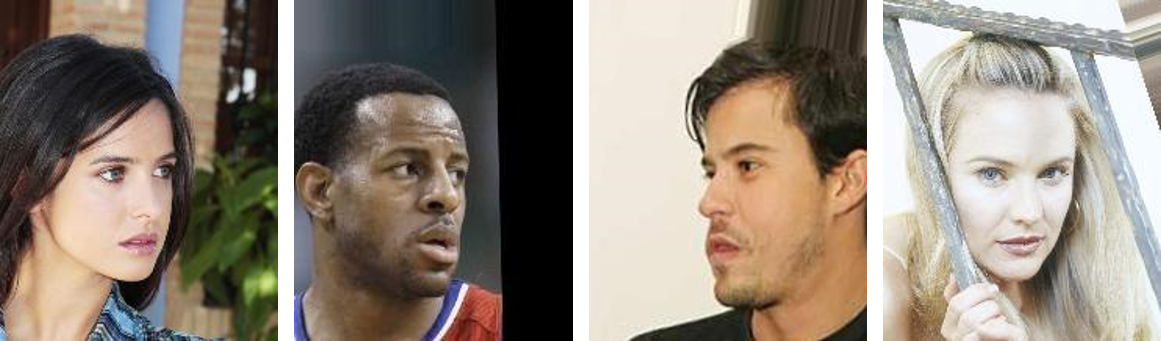}
\caption{Sample images of CelebA correctly classified only by the proposed supervised COSMOS model.}
\label{fig:samples}
\end{center}
\vspace{-15pt}
\end{figure}

\textbf{Effect of Number of Layers, Initialization, and Tessellation:} Table \ref{tab:depth} demonstrates the effect of varying the number of layers of Supervised COSMOS. The best results for CelebA dataset are obtained with seven hidden layers, and a reduction in accuracy is observed as we go further. Similarly, the best results for SVHN dataset are obtained using six hidden layers. Fig. \ref{fig:bar} presents the accuracy obtained by using Supervised COSMOS and Supervised COSMOS with tessellation on SVHN and CelebA datasets. It can be observed that incorporating tessellation improves the performance of the proposed framework by around 2\%. Models learned on the image patches are able to encode local information about the image, while the full image based network focuses more on learning global features. Combining information from both the components results in a holistic feature representation, which in turn enhances the classification performance. Fig. \ref{fig:samples} presents some images of CelebA dataset which were correctly classified by the proposed Supervised COSMOS model only. Upon observing these samples closely, we see that the proposed model handles pose as well as illumination variations, and learns features robust to such variations. These samples further demonstrate the efficacy of the proposed model and motivate its use for robust feature extraction for classification.

\section{Conclusion}
Over the past decade, researchers have actively pursued the domain of deep learning in order to learn robust features and effective classifiers. Deep learning has shown to perform well on several tasks, however, majority of the research has focused on the specific paradigm of Convolutional Neural Networks. While CNNs have been well studied, it is our belief that research along other different paradigms should also be pursued in order to develop competitive algorithms. Another promising paradigm of deep learning is the autoencoder, which learns representative features of the input. In this research we have developed a novel autoencoder formulation, termed as the Supervised COSMOS autoencoder, which learns features specifically for the task of classification. The proposed autoencoder has a multi-objective loss function that incorporates (i) Cosine similarity to encode ``direction'' information, (ii) Mahalanobis distance to encode ``distribution'' information of the input with respect to the reconstruction, and (iii) Mutual Information based supervision in order to learn discriminative features. This enables the model to learn \textit{supervised} features invariant to minor vector variations in illumination and rotation. Experimental evaluations on image classification, attribute prediction, and face recognition showcase the versatility of the proposed approach. State-of-the-art results are obtained on standard benchmark datasets such as MNIST, CIFAR-10, SVHN, CelebA, LFWA, Adience, and IJB-A, which demonstrate the effectiveness of the proposed Supervised COSMOS autoencoder.

\section{Acknowledgments}
This research is partly supported by MEITY (Govt. of India). M. Singh, M. Vatsa, and R. Singh are partly supported by Infosys CAI at IIIT-Delhi, and S.Nagpal is partly supported via TCS PhD Fellowship.  

\bibliographystyle{IEEEtran}
\bibliography{bibFile}

\begin{thebibliography}{10}
\providecommand{\url}[1]{#1}
\csname url@samestyle\endcsname
\providecommand{\newblock}{\relax}
\providecommand{\bibinfo}[2]{#2}
\providecommand{\BIBentrySTDinterwordspacing}{\spaceskip=0pt\relax}
\providecommand{\BIBentryALTinterwordstretchfactor}{4}
\providecommand{\BIBentryALTinterwordspacing}{\spaceskip=\fontdimen2\font plus
\BIBentryALTinterwordstretchfactor\fontdimen3\font minus
  \fontdimen4\font\relax}
\providecommand{\BIBforeignlanguage}[2]{{%
\expandafter\ifx\csname l@#1\endcsname\relax
\typeout{** WARNING: IEEEtran.bst: No hyphenation pattern has been}%
\typeout{** loaded for the language `#1'. Using the pattern for}%
\typeout{** the default language instead.}%
\else
\language=\csname l@#1\endcsname
\fi
#2}}
\providecommand{\BIBdecl}{\relax}
\BIBdecl

\bibitem{lecun2015deep}
Y.~LeCun, Y.~Bengio, and G.~Hinton, ``Deep learning,'' \emph{Nature}, vol. 521,
  no. 7553, 2015.

\bibitem{ae}
G.~Hinton and R.~Salakhutdinov, ``Reducing the dimensionality of data with
  neural networks,'' \emph{Science}, vol. 313, no. 5786, pp. 504 -- 507, 2006.

\bibitem{majumdar2016}
A.~Majumdar, R.~Singh, and M.~Vatsa, ``Face recognition via class sparsity
  based supervised encoding,'' \emph{IEEE Transactions on Pattern Analysis and
  Machine Intelligence}, vol.~39, pp. 1273--1280, 2017.

\bibitem{xu16}
J.~Xu, L.~Xiang, Q.~Liu, H.~Gilmore, J.~Wu, J.~Tang, and A.~Madabhushi,
  ``Stacked sparse autoencoder {(SSAE)} for nuclei detection on breast cancer
  histopathology images,'' \emph{IEEE Transactions on Medical Imaging},
  vol.~35, no.~1, pp. 119--130, 2016.

\bibitem{varAE}
W.~Xu, H.~Sun, C.~Deng, and Y.~Tan, ``Variational autoencoder for
  semi-supervised text classification,'' in \emph{{AAAI} Conference on
  Artificial Intelligence}, 2017, pp. 3358--3364.

\bibitem{aeAAAI16}
S.~Zhai and Z.~M. Zhang, ``Semisupervised autoencoder for sentiment analysis,''
  in \emph{AAAI Conference on Artificial Intelligence}, 2016, pp. 1394--1400.

\bibitem{zhang14}
J.~Zhang, S.~Shan, M.~Kan, and X.~Chen, ``Coarse-to-fine auto-encoder networks
  {(CFAN)} for real-time face alignment,'' in \emph{European Conference on
  Computer Vision}, 2014, pp. 1--16.

\bibitem{sparse}
A.~Ng, ``Sparse autoencoder,'' \emph{CS294A Lecture notes}, vol.~72, pp. 1--19,
  2011.

\bibitem{sdae}
P.~Vincent, H.~Larochelle, I.~Lajoie, Y.~Bengio, and P.~A. Manzagol, ``Stacked
  denoising autoencoders: Learning useful representations in a deep network
  with a local denoising criterion,'' \emph{Journal of Machine Learning
  Research}, vol.~11, pp. 3371--3408, 2010.

\bibitem{highContract}
S.~Rifai, G.~Mesnil, P.~Vincent, X.~Muller, Y.~Bengio, Y.~Dauphin, and
  X.~Glorot, ``Higher order contractive auto-encoder,'' in \emph{European
  Conference on Machine Learning and Knowledge Discovery in Databases}, 2011,
  pp. 645--660.

\bibitem{Contractive}
S.~Rifai, P.~Vincent, X.~Muller, X.~Glorot, and Y.~Bengio, ``Contractive
  auto-encoders: Explicit invariance during feature extraction,'' in
  \emph{International Conference on Machine Learning}, 2011, pp. 833--840.

\bibitem{transformingAE}
G.~E. Hinton, A.~Krizhevsky, and S.~D. Wang, ``Transforming auto-encoders,'' in
  \emph{International Conference on Artificial Neural Networks}, 2011, pp.
  44--51.

\bibitem{generalizedAE}
W.~Wang, Y.~Huang, Y.~Wang, and L.~Wang, ``Generalized autoencoder: A neural
  network framework for dimensionality reduction,'' in \emph{IEEE Conference on
  Computer Vision and Pattern Recognition Workshops}, 2014, pp. 496--503.

\bibitem{vae}
D.~P. Kingma and M.~Welling, ``Stochastic gradient vb and the variational
  auto-encoder,'' in \emph{International Conference on Learning
  Representations}, 2014.

\bibitem{smcae}
X.~Zhang, Y.~Fu, S.~Jiang, L.~Sigal, and G.~Agam, ``Learning from synthetic
  data using a stacked multichannel autoencoder,'' in \emph{International
  Conference on Machine Learning and Applications}, 2015, pp. 461--464.

\bibitem{gao2015}
S.~Gao, Y.~Zhang, K.~Jia, J.~Lu, and Y.~Zhang, ``Single sample face recognition
  via learning deep supervised autoencoders,'' \emph{IEEE Transactions on
  Information Forensics and Security}, vol.~10, pp. 2108--2118, 2015.

\bibitem{transferAE}
F.~Zhuang, X.~Cheng, P.~Luo, S.~J. Pan, and Q.~He, ``Supervised representation
  learning: Transfer learning with deep autoencoders,'' in \emph{International
  Joint Conference on Artificial Intelligence}, 2015, pp. 4119--4125.

\bibitem{mtae}
M.~Ghifary, W.~Bastiaan~Kleijn, M.~Zhang, and D.~Balduzzi, ``Domain
  generalization for object recognition with multi-task autoencoders,'' in
  \emph{IEEE International Conference on Computer Vision}, 2015, pp.
  2551--2559.

\bibitem{relational}
Q.~Meng, D.~Catchpoole, D.~Skillicom, and P.~J. Kennedy, ``Relational
  autoencoder for feature extraction,'' in \emph{International Joint Conference
  on Neural Networks}, 2017, pp. 364--371.

\bibitem{aeAAAI}
S.~Wang, Z.~Ding, and Y.~Fu, ``Feature selection guided auto-encoder,'' in
  \emph{AAAI Conference on Artificial Intelligence}, 2017.

\bibitem{conditional}
Z.~Zhang, Y.~Song, and H.~Qi, ``Age progression/regression by conditional
  adversarial autoencoder,'' in \emph{IEEE Conference on Computer Vision and
  Pattern Recognition}, 2017.

\bibitem{cra}
L.~Tran, X.~Liu, J.~Zhou, and R.~Jin, ``Missing modalities imputation via
  cascaded residual autoencoder,'' in \emph{IEEE Conference on Computer Vision
  and Pattern Recognition}, 2017.

\bibitem{rcodean}
A.~Sethi, M.~Singh, R.~Singh, and M.~Vatsa, ``Residual codean autoencoder for
  facial attribute analysis,'' \emph{Pattern Recognition Letters}, 2018, doi =
  {https://doi.org/10.1016/j.patrec.2018.03.010}.

\bibitem{coupled}
K.~Zeng, J.~Yu, R.~Wang, C.~Li, and D.~Tao, ``Coupled deep autoencoder for
  single image super-resolution,'' \emph{IEEE transactions on cybernetics},
  vol.~47, no.~1, pp. 27--37, 2017.

\bibitem{semantic}
E.~Kodirov, T.~Xiang, and S.~Gong, ``Semantic autoencoder for zero-shot
  learning,'' in \emph{The IEEE Conference on Computer Vision and Pattern
  Recognition}, July 2017.

\bibitem{anush17}
A.~Sankaran, M.~Vatsa, R.~Singh, and A.~Majumdar, ``Group sparse autoencoder,''
  \emph{Image and Vision Computing}, vol.~60, pp. 64 -- 74, 2017.

\bibitem{contrast}
X.~Zheng, Z.~Wu, H.~Meng, and L.~Cai, ``Contrastive auto-encoder for phoneme
  recognition,'' in \emph{International Conference on Acoustics, Speech and
  Signal Processing}, 2014, pp. 2529--2533.

\bibitem{singh17}
M.~Singh, S.~Nagpal, R.~Singh, and M.~Vatsa, ``Class representative autoencoder
  for low resolution multi-spectral gender classification,'' in
  \emph{International Joint Conference on Neural Networks}, 2017, pp.
  1026--1033.

\bibitem{larsen16}
A.~B.~L. Larsen, S.~K. S{\o}nderby, H.~Larochelle, and O.~Winther,
  ``Autoencoding beyond pixels using a learned similarity metric,'' in
  \emph{International Conference on Machine Learning}, 2016.

\bibitem{mi}
J.~J.-Y. Wang, Y.~Wang, S.~Zhao, and X.~Gao, ``Maximum mutual information
  regularized classification,'' \emph{Engineering Applications of Artificial
  Intelligence}, vol.~37, pp. 1 -- 8, 2015.

\bibitem{am}
W.~Byrne, ``Alternating minimization and {B}oltzmann machine learning,''
  \emph{IEEE Transactions on Neural Networks}, vol.~3, no.~4, pp. 612--620,
  1992.

\bibitem{resNet}
K.~He, X.~Zhang, S.~Ren, and J.~Sun, ``Deep residual learning for image
  recognition,'' in \emph{{IEEE} Conference on Computer Vision and Pattern
  Recognition}, 2016, pp. 770--778.

\bibitem{mnist}
Y.~LeCun and C.~Cortes, ``{MNIST} handwritten digit database,'' \emph{AT\&T
  Labs [Online]. Available: http://yann. lecun. com/exdb/mnist}, 2010.

\bibitem{cifar}
A.~Torralba, R.~Fergus, and W.~T. Freeman, ``80 million tiny images: A large
  data set for nonparametric object and scene recognition,'' \emph{IEEE
  Transactions on Pattern Analysis and Machine Intelligence}, vol.~30, pp.
  1958--1970, 2008.

\bibitem{svhn}
Y.~Netzer, T.~Wang, A.~Coates, A.~Bissacco, B.~Wu, and A.~Y. Ng, ``Reading
  digits in natural images with unsupervised feature learning,'' in \emph{NIPS
  Workshop on Deep Learning and Unsupervised Feature Learning}, 2011.

\bibitem{liu2015deep}
Z.~Liu, P.~Luo, X.~Wang, and X.~Tang, ``Deep learning face attributes in the
  wild,'' in \emph{IEEE International Conference on Computer Vision}, 2015, pp.
  3730--3738.

\bibitem{adience}
E.~Eidinger, R.~Enbar, and T.~Hassner, ``Age and gender estimation of
  unfiltered faces,'' \emph{IEEE Transactions on Information Forensics and
  Security}, vol.~9, no.~12, pp. 2170--2179, 2014.

\bibitem{ijba}
B.~F. Klare, B.~Klein, E.~Taborsky, A.~Blanton, J.~Cheney, K.~Allen,
  P.~Grother, A.~Mah, M.~Burge, and A.~K. Jain, ``Pushing the frontiers of
  unconstrained face detection and recognition: {IARPA} {J}anus {B}enchmark
  {A},'' in \emph{IEEE Conference on Computer Vision and Pattern Recognition},
  2015, pp. 1931--1939.

\bibitem{wtaAE}
A.~Makhzani and B.~J. Frey, ``Winner-take-all autoencoders,'' in \emph{Advances
  in Neural Information Processing Systems}, C.~Cortes, N.~D. Lawrence, D.~D.
  Lee, M.~Sugiyama, and R.~Garnett, Eds., 2015, pp. 2791--2799.

\bibitem{aae}
A.~Makhzani, J.~Shlens, N.~Jaitly, and I.~Goodfellow, ``Adversarial
  autoencoders,'' in \emph{International Conference on Learning
  Representations}, 2016.

\bibitem{spae}
T.~Yu, C.~Guo, L.~Wang, S.~Xiang, and C.~Pan, ``Self-paced autoencoder,''
  \emph{IEEE Signal Processing Letters}, vol.~25, no.~7, pp. 1054--1058, 2018.

\bibitem{dropconnect}
L.~Wan, M.~Zeiler, S.~Zhang, Y.~L. Cun, and R.~Fergus, ``Regularization of
  neural networks using dropconnect,'' in \emph{International Conference on
  Machine Learning}, vol.~28, no.~3, 2013, pp. 1058--1066.

\bibitem{mcdnn}
D.~Ciregan, U.~Meier, and J.~Schmidhuber, ``Multi-column deep neural networks
  for image classification,'' in \emph{IEEE Conference on Computer Vision and
  Pattern Recognition}, 2012, pp. 3642--3649.

\bibitem{genPoolCnn}
C.~Lee, P.~W. Gallagher, and Z.~Tu, ``Generalizing pooling functions in
  convolutional neural networks: Mixed, gated, and tree,'' in
  \emph{International Conference on Artificial Intelligence and Statistics},
  2016, pp. 464--472.

\bibitem{recCnn}
M.~Liang and X.~Hu, ``Recurrent convolutional neural network for object
  recognition,'' in \emph{IEEE Conference on Computer Vision and Pattern
  Recognition}, 2015, pp. 3367--3375.

\bibitem{pieceWise}
Z.~Liao and G.~Carneiro, ``On the importance of normalisation layers in deep
  learning with piecewise linear activation units,'' in \emph{IEEE Winter
  Conference on Applications of Computer Vision}, 2016.

\bibitem{goodInit}
\BIBentryALTinterwordspacing
D.~Mishkin and J.~Matas, ``All you need is a good init,'' \emph{CoRR}, vol.
  abs/1511.06422, 2015. [Online]. Available:
  \url{http://arxiv.org/abs/1511.06422}
\BIBentrySTDinterwordspacing

\bibitem{boDnn}
J.~Snoek, O.~Rippel, K.~Swersky, R.~Kiros, N.~Satish, N.~Sundaram, M.~Patwary,
  M.~Prabhat, and R.~Adams, ``Scalable {B}ayesian optimization using deep
  neural networks,'' in \emph{International Conference on Machine Learning},
  2015, pp. 2171--2180.

\bibitem{wide}
S.~Zagoruyko and N.~Komodakis, ``Wide residual networks,'' in \emph{British
  Machine Vision Conference}, 2016, pp. 87.1--87.12.

\bibitem{denseNet}
G.~Huang, Z.~Liu, K.~Q. Weinberger, and L.~van~der Maaten, ``Densely connected
  convolutional networks,'' in \emph{IEEE Conference on Computer Vision and
  Pattern Recognition}, 2017.

\bibitem{capsule}
S.~Sabour, N.~Frosst, and G.~E. Hinton, ``Dynamic routing between capsules,''
  in \emph{Advances in Neural Information Processing Systems}, 2017, pp.
  3856--3866.

\bibitem{houVAE}
X.~Hou, L.~Shen, K.~Sun, and G.~Qiu, ``Deep feature consistent variational
  autoencoder,'' in \emph{{IEEE Winter Conference on Applications of Computer
  Vision}}, 2017, pp. 1133--1141.

\bibitem{wang2016walk}
J.~Wang, Y.~Cheng, and R.~S. Feris, ``Walk and learn: Facial attribute
  representation learning from egocentric video and contextual data,'' in
  \emph{IEEE Conference on Computer Vision and Pattern Recognition}, 2016, pp.
  2295--2304.

\bibitem{zhong2016leveraging}
Y.~Zhong, J.~Sullivan, and H.~Li, ``Leveraging mid-level deep representations
  for predicting face attributes in the wild,'' in \emph{IEEE International
  Conference on Image Processing}, 2016, pp. 3239--3243.

\bibitem{rozsa2016facial}
A.~Rozsa, E.~M. Rudd, and T.~E. Boult, ``Adversarial diversity and hard
  positive generation,'' in \emph{IEEE Conference on Computer Vision and
  Pattern Recognition Workshops}, 2016, pp. 25--32.

\bibitem{rudd2016moon}
E.~M. Rudd, M.~G{\"u}nther, and T.~E. Boult, ``Moon: A mixed objective
  optimization network for the recognition of facial attributes,'' in
  \emph{European Conference on Computer Vision}, 2016, pp. 19--35.

\bibitem{hand17aaai}
E.~Hand and R.~Chellappa, ``Attributes for improved attributes: A multi-task
  network utilizing implicit and explicit relationships for facial attribute
  classification,'' in \emph{AAAI Conference on Artificial Intelligence}, 2017.

\bibitem{kalayeh_17_cvpr}
M.~M. Kalayeh, B.~Gong, and M.~Shah, ``Improving facial attribute prediction
  using semantic segmentation,'' in \emph{IEEE Conference on Computer Vision
  and Pattern Recognition}, 2017.

\bibitem{ijcai18He}
K.~He, Y.~Fu, W.~Zhang, C.~Wang, Y.-G. Jiang, F.~Huang, and X.~Xue,
  ``Harnessing synthesized abstraction images to improve facial attribute
  recognition,'' in \emph{International Joint Conference on Artificial
  Intelligence}, 2018, pp. 733--740.

\bibitem{wang17}
F.~Wang, H.~Han, S.~Shan, and X.~Chen, ``Deep multi-task learning for joint
  prediction of heterogeneous face attributes,'' in \emph{IEEE International
  Conference on Automatic Face Gesture Recognition}, 2017, pp. 173--179.

\bibitem{han17}
H.~Han, A.~K. Jain, S.~Shan, and X.~Chen, ``Heterogeneous face attribute
  estimation: A deep multi-task learning approach,'' \emph{IEEE Transactions on
  Pattern Analysis and Machine Intelligence}, 2018,
  doi={10.1109/TPAMI.2017.2738004}.

\bibitem{leviCvpr}
G.~Levi and T.~Hassner, ``Age and gender classification using convolutional
  neural networks,'' in \emph{IEEE Conference on Computer Vision and Pattern
  Recognition Workshops}, 2015, pp. 34--42.

\bibitem{dex}
R.~Rothe, R.~Timofte, and L.~J.~V. Gool, ``{DEX:} deep expectation of apparent
  age from a single image,'' in \emph{{IEEE} International Conference on
  Computer Vision Workshop}, 2015, pp. 252--257.

\bibitem{cnnSvm}
J.~v.~d. Wolfshaar, M.~F. Karaaba, and M.~A. Wiering, ``Deep convolutional
  neural networks and support vector machines for gender recognition,'' in
  \emph{IEEE Symposium Series on Computational Intelligence}, 2015, pp.
  188--195.

\bibitem{transfer}
G.~Ozbulak, Y.~Aytar, and H.~K. Ekenel, ``How transferable are {CNN}-based
  features for age and gender classification?'' in \emph{International
  Conference of the Biometrics Special Interest Group}, 2016.

\bibitem{fdar}
H.~Liu, X.~Shen, and H.~Ren, ``{FDAR-N}et: Joint convolutional neural networks
  for face detection and attribute recognition,'' in \emph{International
  Symposium on Computational Intelligence and Design}, vol.~2, 2016, pp.
  184--187.

\bibitem{pr17}
P.~Rodríguez, G.~Cucurull, J.~M. Gonfaus, F.~X. Roca, and J.~Gonzàlez, ``Age
  and gender recognition in the wild with deep attention,'' \emph{Pattern
  Recognition}, vol.~72, pp. 563 -- 571, 2017.

\bibitem{dcnn}
J.~C. Chen, R.~Ranjan, A.~Kumar, C.~H. Chen, V.~M. Patel, and R.~Chellappa,
  ``An end-to-end system for unconstrained face verification with deep
  convolutional neural networks,'' in \emph{IEEE International Conference on
  Computer Vision Workshop}, 2015, pp. 360--368.

\bibitem{hua-2016}
J.~Yang, P.~Ren, D.~Chen, F.~Wen, H.~Li, and G.~Hua, ``Neural aggregation
  network for video face recognition,'' \emph{CoRR}, vol. abs/1603.05474, 2016.

\bibitem{tdff-tpe}
\BIBentryALTinterwordspacing
L.~Xiong, J.~Karlekar, J.~Zhao, J.~Feng, S.~Pranata, and S.~Shen, ``A good
  practice towards top performance of face recognition: Transferred deep
  feature fusion,'' \emph{CoRR}, vol. abs/1704.00438, 2017. [Online].
  Available: \url{http://arxiv.org/abs/1704.00438}
\BIBentrySTDinterwordspacing

\bibitem{chellappa-l2}
\BIBentryALTinterwordspacing
R.~Ranjan, C.~D. Castillo, and R.~Chellappa, ``L2-constrained softmax loss for
  discriminative face verification,'' \emph{CoRR}, vol. abs/1703.09507, 2017.
  [Online]. Available: \url{http://arxiv.org/abs/1703.09507}
\BIBentrySTDinterwordspacing

\bibitem{vggface2}
Q.~Cao, L.~Shen, W.~Xie, O.~M. Parkhi, and A.~Zisserman, ``Vggface2: A dataset
  for recognising faces across pose and age,'' in \emph{IEEE International
  Conference on Automatic Face Gesture Recognition}, 2018, pp. 67--74.

\end{thebibliography}

\ifCLASSOPTIONcaptionsoff
  \newpage
\fi

\end{document}